\documentclass[10pt,twocolumn,letterpaper]{article}

\usepackage{amsmath,amsfonts,bm}

\def\eqref#1{equation~\ref{#1}}

\def\1{\bm{1}}
\newcommand{\train}{\mathcal{D}_{trans}}
\newcommand{\pretrain}{\mathcal{D_\mathrm{pre}}}

\newcommand{\test}{\mathcal{D}_\mathrm{test}}

\DeclareMathAlphabet{\mathsfit}{\encodingdefault}{\sfdefault}{m}{sl}
\SetMathAlphabet{\mathsfit}{bold}{\encodingdefault}{\sfdefault}{bx}{n}

\def\gL{{\mathcal{L}}}

\def\gX{{\mathcal{X}}}
\def\gY{{\mathcal{Y}}}
\def\gZ{{\mathcal{Z}}}

\newcommand{\E}{\mathbb{E}}

\usepackage[pagenumbers]{cvpr} 

\usepackage{graphicx}
\usepackage{amsmath}
\usepackage{amssymb}
\usepackage{booktabs}

\usepackage[pagebackref,breaklinks,colorlinks]{hyperref}

\usepackage[capitalize]{cleveref}
\crefname{section}{Sec.}{Secs.}
\Crefname{section}{Section}{Sections}
\Crefname{table}{Table}{Tables}
\crefname{table}{Tab.}{Tabs.}

\def\cvprPaperID{3535} 
\def\confName{CVPR}
\def\confYear{2022}

\newcommand{\risk}{R}
\newcommand{\pr}{\bm{\SR_f \hyphen \SR_\mathrm{scratch}}}
\newcommand{\lcurve}{\bm{\SR_f \hyphen n}}

\newcommand{\SR}{\mathrm{cR}}
\mathchardef\hyphen="2D
\begin{document}

\title{
Simple Control Baselines for Evaluating Transfer Learning
\vspace{-4pt}
}


\author{
Andrei Atanov$^{\dagger*}$, Shijian Xu$^{\dagger}$\thanks{equal contribution} , Onur Beker$^{\dagger}$, Andrei Filatov$^{\dagger\S}$, Amir Zamir$^{\dagger}$\vspace{7pt}\\
	\hspace{-2pt}$^\dagger$\hspace{-2pt} Swiss Federal Institute of Technology (EPFL)\;\;  
	$^\S$\hspace{-2pt}  Moscow Institute of Physics and Technology (MIPT) \vspace{7pt}\\
	\textcolor{blue}{\url{https://transfer-controls.epfl.ch}\vspace{-9pt}}
}


\maketitle

\begin{abstract}
Transfer learning has witnessed remarkable progress in recent years, for example, with the introduction of augmentation-based contrastive self-supervised learning methods. While a number of large-scale empirical studies on the transfer performance of such models have been conducted, there is not yet an agreed-upon set of control baselines, evaluation practices, and metrics to report, which often hinders a nuanced and calibrated understanding of the real efficacy of the methods. We share an evaluation standard that aims to quantify and communicate transfer learning performance in an informative and accessible setup. This is done by baking a number of simple yet critical control baselines in the evaluation method, particularly the `blind-guess' (quantifying the dataset bias), `scratch-model' (quantifying the architectural contribution), and `maximal-supervision' (quantifying the upper-bound). To demonstrate how the evaluation standard can be employed, we provide an example empirical study investigating a few basic questions about self-supervised learning. For example, using this standard, the study shows the effectiveness of existing self-supervised pre-training methods is skewed towards image classification tasks versus dense pixel-wise predictions. In general, we encourage using/reporting the suggested control baselines in evaluating transfer learning in order to gain a more meaningful and informative understanding.

\end{abstract}

\section{Introduction}


Creating computer vision models that can support a wide range of downstream tasks will require the development of representations whose utility extend beyond the exact same objective they were optimized for. 
Transfer learning is a general approach to operationalize this perspective by using representations of a model trained on a source task to improve the sample complexity of learning another downstream task. 
Supervised pre-training approaches to transfer learning, however, still have the drawback of requiring large annotated datasets for the source task.
Self-supervised learning, in turn, holds the potential to remove this bottleneck by learning representations directly from unlabelled data by means of employing a proxy pre-training objective whose optimization results in transferable representations, but does not require manual labelling \cite{noroozi2016unsupervised, zhang2016colorful, gidaris2018unsupervised}. For example, the recently revisited contrastive learning approaches \cite{hadsell2006dimensionality, chen2020simple, he2020momentum, caron2020unsupervised} constitute a milestone in this paradigm and were shown to achieve impressive transfer performance competitive with representations obtained in a fully supervised way for classification tasks. 

\begin{figure}[t]
    \centering
    \includegraphics[width=\linewidth]{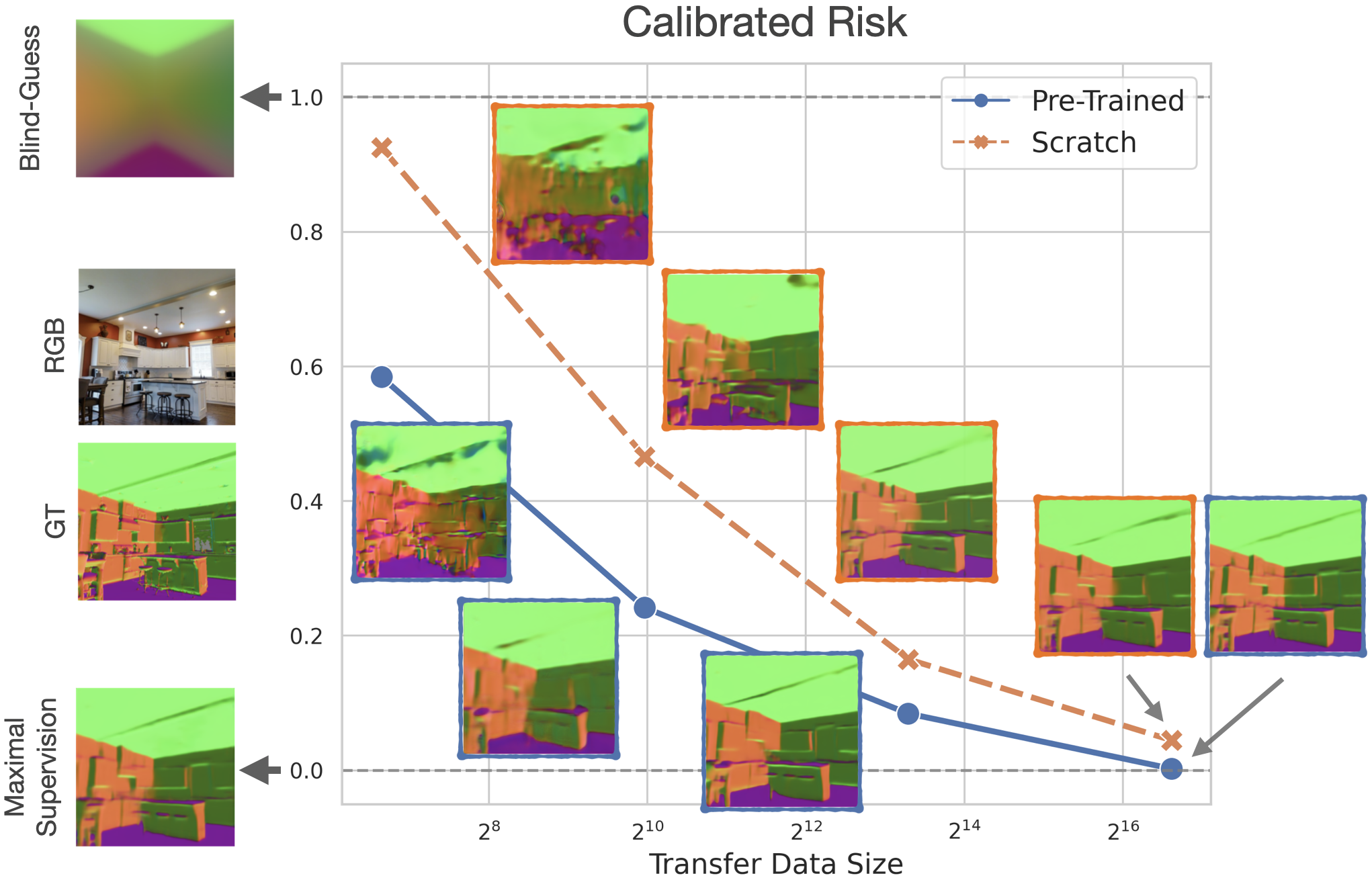}
    \vspace*{-5mm}
    \caption{
    \textbf{What are the most informative control baselines and transfer dataset sizes that provide sufficient context for evaluating the effectiveness of a pre-training method?}
    We employ a calibrated evaluation standard that incorporates: 1) a \textit{scratch control baseline} to account for the architectural inductive biases, 
    2) a \textit{maximal-supervision baseline} to approximate the best achievable performance,
    and 3) a \textit{blind-guess baseline} to account for the dataset bias.
    We consider a range of data sizes across which the performance of the scratch baseline varies between the blind-guess level (where there is large room for improvement; plot's left) and maximal-supervision level (where nearly no pre-training is needed; plot's right). The figure illustrates the performance of an example model, where the x and y axes show the data size and calibrated performance, respectively, along with qualitative instances along the curves.
    }
    \vspace*{-6mm}
    \label{fig:fig_pull}
\end{figure}

Given the rapid progress in transfer and self-supervised learning, 
obtaining a clear and complete picture of the increasingly large number of methods being proposed is difficult, in part due to an absence of standard evaluation practices and the non-uniformity of reported (control) baselines.
Ideally, such an evaluation standard should have a number of desirable properties, such as: 
\begin{enumerate}
    \item providing a simple metric that clearly communicates the ``effectiveness" transfer learning of a given method on a natural and interpretable scale.
    \item taking into account the effects of \textit{irreducible factors that upper-bound the best possible transfer performance}, such as the uncertainty in observations or the imperfect nature of current neural network training practices.
    \item normalizing away the benefits due to the statistical regularities of datasets that \textit{lower-bound the worst possible performance} \cite{coughlan2000manhattan}, to prevent them from being a confounder in comparisons.
    \item providing an intuitive way to draw comparisons about the utility of different self-supervised learning methods \textit{across different downstream tasks}.
\end{enumerate}

This paper proposes an evaluation standard that satisfies the aforementioned properties by incorporating a set of key control baselines in the evaluation. We then proceed with example analyses on questions that would be otherwise harder to study without such a standard -- for example, assessing whether existing self-supervised learning methods are equally effective across different types of tasks. The proposed standard is not limited to self-supervised learning and is applicable to any transfer learning evaluation.

\section{Related Work}
\textbf{Self-Supervised Representation Learning} is a special case of general transfer learning and one of the most attended ones today. Earlier work on self-supervised learning largely focused on hand-designed proxy-tasks, such as solving a jigsaw puzzle \cite{noroozi2016unsupervised}, image colorization \cite{zhang2016colorful}, rotation prediction \cite{gidaris2018unsupervised}, as well as numerous others \cite{misra2020self, pathak2016context, zamir2016generic}.
More recently, a number of augmentation-based contrastive methods have been proposed \cite{chen2020simple, he2020momentum, caron2020unsupervised, henaff2020data, chen2020big}, and were shown to outperform previous approaches significantly.
These methods employ a contrastive-loss formulation \cite{hadsell2006dimensionality}, and learn representations that are similar for augmented views of the same image and dissimilar for different images. \cite{bardes2021vicreg, zbontar2021barlow, grill2020bootstrap} also suggest learning representations that are close for augmentations of the same image, but do not use negative examples explicitly and achieve similar performance to contrastive methods.



\textbf{Benchmarking and Analysis.} The literature contains various empirical studies on the effectiveness of self-supervised pre-training, across a wide range of methods and downstream tasks.
 \cite{van2021benchmarking, islam2021broad, cole2021does} focus on classification tasks and study how the final transfer performance depends on the image domain of a dataset, downstream-task complexity, and the available amount of labelled data for transfer.
\cite{ericsson2021well, newell2020useful} further include non-classification tasks in their evaluation setup, and \cite{kotar2021contrasting} conducts a comprehensive empirical study with a diverse set of downstream tasks to demonstrate the benefits of contrastive self-supervised pre-training.
 \cite{goyal2019scaling, kolesnikov2019revisiting, tian2020makes, xiao2020should, cole2021does} examine the effects of various pre-training parameters such as the model capacity, pretext task complexity, and the augmentation policy used in the contrastive loss formulation, to provide a further understanding of the successes of self-supervised pre-training.
\cite{tian2020makes, tsai2020self, arora2019theoretical, lee2020predicting} build a mathematical framework to explain how self-supervised pre-training can improve the performance on downstream tasks, aiming to provide theoretical guarantees.
In this work, we also include non-classification tasks in our showcase analysis, and aim to study the transfer performance more comprehensively with the proposed control baselines and evaluations across different data-regimes.

\textbf{Evaluation Metrics and Baselines.} A standard way to evaluate self-supervised learning methods is through freezing their pre-trained representations and reporting the performance of a linear classifier trained on top to solve ImageNet classification.
\cite{ericsson2021well} show empirically that this practice is not as predictive of the downstream performance for non-classification tasks as it is for classification tasks.
\cite{kotar2021contrasting} adopt a more comprehensive setup that evaluates transfer performance on a broader range of downstream tasks, and opt for reporting most of their results in terms of relative improvements over ImageNet supervised pre-training, but the absolute level of performance of the ImageNet pre-training baseline remains unclear for their study.
\cite{newell2020useful} and  \cite{cole2021does} include comparisons to a model trained from scratch on the same amount of labelled data, and the former further suggests measuring the transfer efficacy as the proportion of labelled data saved by self-supervised pre-training compared to training a model from scratch to achieve the same performance.
Our work also incorporates the scratch performance as a baseline to account for the difficulty of the downstream task.
We additionally put our comparisons into perspective by attempting to account for the dataset bias and the performance upper-bound due to irreducible uncertainties or architectural limitations.

\section{The Evaluation Standard}

\label{sec:metric}
This section describes the control baselines and the overall evaluation standard we propose. We start with a short description of the transfer-learning setting, and then proceed with presenting the definitions and motivations behind each of the control baselines. We then incorporate these baselines into an affine rescaling of the empirical risk, and discuss the associated visualizations and additional metrics.
\subsection{Transfer Learning Setting}
\label{sec:transfer}
We consider a standard transfer learning setting that employs an encoder-decoder architecture.
The \textit{encoder} $\psi: \gX \to \gZ$, where $\gX$ and $\gZ$ denote the set of input images and their projected latent representations respectively, is first optimized using a given \textit{pre-training method } (e.g., a self-supervised method, or a source-task solved in a fully supervised way) on a \textit{pre-training dataset} $\pretrain$. 
The \textit{decoder} $\phi: \gZ \to \gY$ for the \textit{downstream task}, with $\gY$ denoting the output space, is then trained in a supervised way using a \textit{transfer dataset} $\train$ of annotated images to minimize the task-specific loss $\gL$, and the encoder parameters are also fine-tuned. The number of images $|\train|$ used for this optimization is called the \textit{data-regime} for transfer learning. 
The resulting model is denoted as $f_\theta = \phi \circ \psi$, where $\theta$ includes all the network parameters. 
After training, the empirical risk is computed over the test partition $\test$ of the transfer dataset using the loss $\gL$ for the downstream-task:
\begin{equation}
    \risk_{f_\theta} = \dfrac{1}{|\test|} \sum_{(x, y) \in \test} \gL(y, f_\theta(x))).
\end{equation}
\subsection{Control Baselines}
\label{sec:transfer}
\textbf{Motivation}. Transfer learning results are commonly reported using the empirical risk. Therefore, interpreting such results requires knowledge about the scale of the training loss and its nonlinear relation to the inherent prediction fidelity/value for the particular downstream task and dataset. To illustrate with a toy example, when a classification model is reported to achieve the accuracy of $0.9$, judging whether this model performs well or not cannot be done solely by means of the reported metric: the model can be considered proficient if there are 1000 uniformly distributed classes in the dataset, but not so if $90\%$ of the images belong to the same class -- since the accuracy of merely statistically informed guess is 0.001 (900$\times$ worse than the model's) for the former and 0.9 (the same as the model's) for the latter. Furthermore, the sample complexity of obtaining an \emph{additional improvement} will differ depending on the granularity and similarity of the categories.

Such observations motivate the following questions: What additional context can we provide to create a more complete picture of the transfer performance?
What are the most sensible, efficient, and simple control baselines that would allow us to attain a more accurate assessment?

\begin{figure*}
    \vspace*{-6mm}
    \centering
    \includegraphics[width=\textwidth]{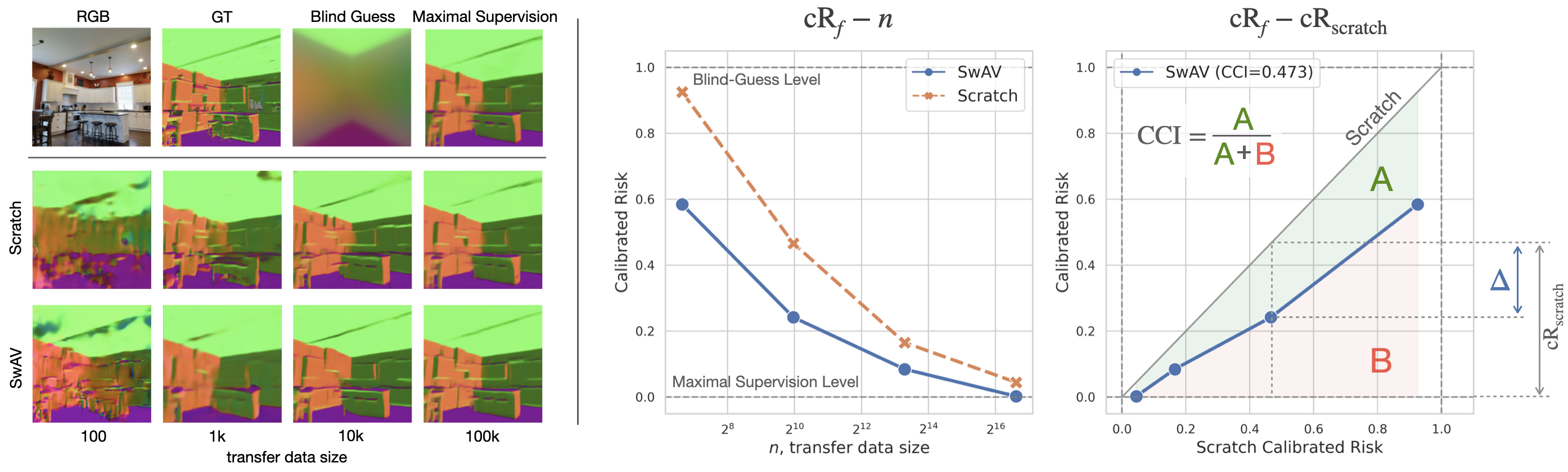}
    \vspace*{-7mm}
    \caption{
    Right: Exemplary $\lcurve$ and $\pr$ curves for the SwAV self-supervised pre-training method, transferred to surface normals estimation. The relative improvement against the scratch control baseline (i.e. $\Delta/\SR_\mathrm{scratch}$) can directly be read from the plot and provides a visualization for the transfer efficacy of a given method.
    Left: corresponding qualitative surface normals predictions for SwAV and three control baselines: blind-guess, maximal-supervision and scratch.
    In this example, the improvement of the SSL pre-training over scratch is observed to diminishes with more training data, which can also be inferred from the qualitative examples for 10K/100K. This (rather general) trend indicates the benefit of transfer learning in high-data regime is less significant, and therefore, relative improvement comparisons are more meaningful across different data-regimes (particularly mid and low data) where the scratch baseline performs relatively poorly.
    }
    \label{fig:fig1}
    \vspace*{-6mm}
\end{figure*}

\textbf{The Scratch Control Baseline:} When training a decoder $\phi$ on a transfer dataset, choices about training parameters (e.g., the model architecture or the optimization method used) are an important source of inductive bias.
Similarly, those specifics of a downstream task and dataset that characterize their amenability for model training have an influence on how the optimization will proceed.
Such considerations are independent of the particular choice of pre-training method being evaluated, yet they have a critical impact on the final transfer performance. 
Therefore, as previous studies also illustrate \cite{cole2021does}, it is important to include the \textit{scratch control baseline} (i.e., a randomly initialized model trained only using a transfer dataset corresponding to the data-regime being examined), to disentangle and clarify the benefits of the pre-training method.

\textbf{The Maximal-Supervision Control Baseline:} Even in the presence of an infinite amount of data, many factors can still limit the minimum error achievable on a downstream task. Examples include uncertainties and multi-modality in the label generation process, finite model capacities and architectural bottlenecks, or the non-convexity of the optimization problem and the imperfect nature of the algorithms employed. Having an estimate of such factors is useful for framing the transfer performance in an interpretable scale.
We, therefore, employ a \textit{maximal-supervision control baseline} -- a randomly initialized model trained using a large amount of data for the downstream task intended to provide \emph{an approximation} to the maximum achievable performance on the downstream task.
This baseline may not always be attainable, e.g., when sufficiently large training data for the dataset domain or a theoretical approximation are unavailable. 
In our experiments, we will show cases where a good approximation is attainable (e.g., for Tasknomomy~\cite{zamir2018taskonomy}) as well as cases where it is not attainable (e.g., for CIFAR-100~\cite{krizhevsky2009learning}), and discuss the implications.

\textbf{The Blind-Guess Control Baseline:} 
Regularities in datasets can allow statistically informed guesses for the downstream task \cite{coughlan2000manhattan}. Such guesses can often be unexpectedly performant~\cite{zamir2018taskonomy,eftekhar2021omnidata}.
The effects of such regularities can be accounted for by constructing an input-agnostic \textit{blind-guess control baseline}, defined as the single constant prediction with the lowest error $\arg\min_{\hat{y} \in \gY} \E_y\left[\gL(\hat{y}, y)\right]$, which corresponds to the mean output for $L_2$ loss, the median output for $L_1$ loss, and the most-represented class for the 0-1 loss.
This control baseline serves as a sanity check to clarify whether a pre-training method provides a tangible benefit, as compared to solely capturing regularities in dataset statistics in an input-independent prediction.

\textbf{Importance of Control Baselines:} 
The main goal of incorporating control baselines is to project the performance of a method onto a comparative scale to capture \textit{how well it works}, as well as to provide insight into \textit{why it works} through rejecting a set of null-hypotheses. Computer vision literature contains numerous recent examples of how the inclusion of appropriate baselines can provide such a clarification for their respective fields and problems. For example, Tatarchenko et al.\cite{tatarchenko2019single} employs a set of recognition baselines for the single-view 3D reconstruction problem, and reveals that the performance of state of the art models are statistically indistinguishable from classification and retrieval based methods. Another example is the study by Zamir et al.\cite{zamir2018taskonomy}, which employs a gain metric (i.e., the win rate against a network trained from scratch) to show that given a target task (e.g., classification), multiple source tasks (e.g., colorization) can collapse to a performance level below the scratch baseline in the lower data regime. Sax et al.\cite{sax2019learning} employs a \textit{blind} intelligent actor baseline (i.e., a policy operating without visual input) for navigation tasks, and reveals that policies trained from scratch as well as state of the art representation learning methods perform at a similar level to this baseline when tested on unseen environments. Here we use such baselines in a more systematic way.

Transfer learning is a field where empirical studies are prevalent, and this empirical nature makes rendering evaluations with respect to an appropriate set of baselines particularly important for measuring the progress in a more accessible and comparable way, while also systematically revealing and preventing blind spots. The current common practice is to report and compare downstream performances between different pre-training methods using the original loss scale or relative to the ImageNet features performance \cite{kotar2021contrasting, ericsson2021well, islam2021broad}.
While this setup still allows conclusions based on quantitative results, it is hard to assess whether the difference in the methods' performance is significant or not.
As an example, comparing the $L_1$ loss for the downstream task of depth estimation as reported in Fig.7 of \cite{kotar2021contrasting} with the blind baseline reported by \cite{zamir2020robust} in Tab.~1, one can conclude that all models perform at the blind guess level, and no meaningful improvement was made. Employing an appropriate set of common control baselines would make it easier to identify and prevent miscalibrations across different studies and gain a more complete and comparable picture about the real effectiveness of methods.

\subsection{The Proposed Evaluation Setup}
\label{sec:calibrated-risk}

\textbf{Calibrated-Risk:} To incorporate all three of the previously proposed control baselines in a single metric, we calibrate the empirical risk $\risk_f$ of a transfer model $f$ as follows: 
\begin{equation}
\label{eq:the-formula-of-everythin}
    \SR_f = \dfrac{\risk_f - \risk_{\mathrm{max}}}{\risk_{\mathrm{blind}} - \risk_{\mathrm{max}}},
\end{equation}
where $\risk_{\mathrm{blind}}$ and $\risk_{\mathrm{max}}$ correspond to the empirical risks of the blind-guess and maximal-supervision controls. We refer to $\SR_f$ as the \textit{calibrated risk}.
Similarly, the calibrated risk of the scratch control is computed as: 
\begin{equation}
\label{eq:the-formula-of-everythin-else}
    \SR_\mathrm{scratch}= \dfrac{\risk_\mathrm{scratch} - \risk_{\mathrm{max}}}{\risk_{\mathrm{blind}} - \risk_{\mathrm{max}}},
\end{equation}
where $\risk_{scratch}$ denotes the empirical risk of the scratch control baseline. It is useful to note that $\SR_f$ and $\SR_{scratch}$ are invariant to affine transformations of the training loss $\gL$. 

The calibrated risk of a given pre-training method varies depending on the \textit{data-regime}, and comparisons across different pre-training methods can therefore show rank reversals. Accordingly, considering the performance \textit{across a range of different data regimes}, in the form of a learning curve \cite{hoiem2021learning}, can provide a more revealing picture of the rigidity and practical value of inductive biases introduced by different pre-training methods. As illustrated in Fig.\ref{fig:fig1}, we suggest two plots to illustrate the resulting curves:

\begin{itemize}
    \item Plotting $\SR_f$ and $\SR_\mathrm{scratch}$ on the y-axis against the transfer dataset size $n$ on the x-axis (i.e., $\lcurve$ curves, as shown in Fig.\ref{fig:fig1}, middle).
    \item Plotting $\SR_f$ on the y-axis against $\SR_\mathrm{scratch}$ on the x-axis, to emphasize the relative improvement against scratch performance (i.e., $\pr$ curves, as shown in Fig.\ref{fig:fig1}, right).
\end{itemize}

\textbf{Reading $\bm{\SR_f \hyphen n}$ Curves:} 
Eq.~\ref{eq:the-formula-of-everythin} implies that calibrated risks for the maximal-supervision and blind-guess control baselines correspond to horizontal lines $\SR_f = 0$ and $\SR_f = 1$. If the maximal-supervision control baseline is trained using sufficiently many data samples (i.e., meaning that $\risk_{\mathrm{max}}$ is a good approximation of the maximum achievable performance, which may not always be true for small-scale datasets), then $\risk_f \leq \risk_{\mathrm{max}}$ holds for all data regimes and per Eq.~\ref{eq:the-formula-of-everythin}, $\SR_f$ approaches the maximal-supervision level $\SR_f = 0$ as $n$ goes to infinity. The $\SR_f$ curve of a good pre-training method should lie below the $\SR_\mathrm{scratch}$ curve and as close to the line $\SR_f = 0$ as possible across all data regimes.

\textbf{Reading $\bm{\SR_f \hyphen \SR_\mathrm{scratch}}$ Curves:}
In most situations, the absolute scalar value $\SR_f$ by itself does not provide a clear indication of how well a pre-training method performs, and the more informative quantity is instead the \textit{relative improvement } with respect to the scratch performance $\SR_\mathrm{scratch}$. 
To visually facilitate this comparison, we propose plotting $\SR_f$ against $\SR_\mathrm{scratch}$.
This maps the $\SR_\mathrm{scratch} \hyphen n$ curve onto the main diagonal $x=y$, and a good pre-training method should therefore lie below it.
The relative-improvement of a pre-training method over scratch can be read from the ratio of distances $\Delta$ and $\SR_\mathrm{scratch}$ in Fig.~\ref{fig:fig1}.

\textbf{The Choice of Transfer Data-Regimes.}
A qualitative assessment of the images given in Fig.~\ref{fig:fig1} and Fig.~\ref{fig:pr-wr-all-tasks} illustrates how the efficiency of self-supervised pre-training compared to the scratch baseline depends on the amount of data available for transfer. Both methods perform similarly in high data-regimes and approach the performance of the maximal-supervision control, which can also be seen from qualitative examples for 100k train sizes.
It is important, therefore, to choose training sizes appropriately when evaluating the efficiency of SSL pre-training.
We define low- and high-data regimes as ones where the scratch performance is close to blind-guess and maximal-supervision controls respectively, and suggest choosing training-dataset sizes to cover this range.

\textbf{Calibrated Cumulative Improvement.}
In order to facilitate tabular analysis and allow straightforward comparisons between different pre-training methods, we introduce a global \textit{Calibrated Cumulative Improvement} (CCI) metric.
We calculate it as the area between an SSL curve and the main diagonal as measured on the $\pr$ plot, divided by the total area under the main diagonal, shown in Fig.~\ref{fig:fig1}.
CCI shows the total improvement of a pre-training method compared to scratch over multiple data-regimes. The CCI values of our experiments are available in the legends of each figure.

\section{Experiment Settings}
\begin{figure*}
    \vspace*{-5mm}
    \centering
    \includegraphics[width=\textwidth]{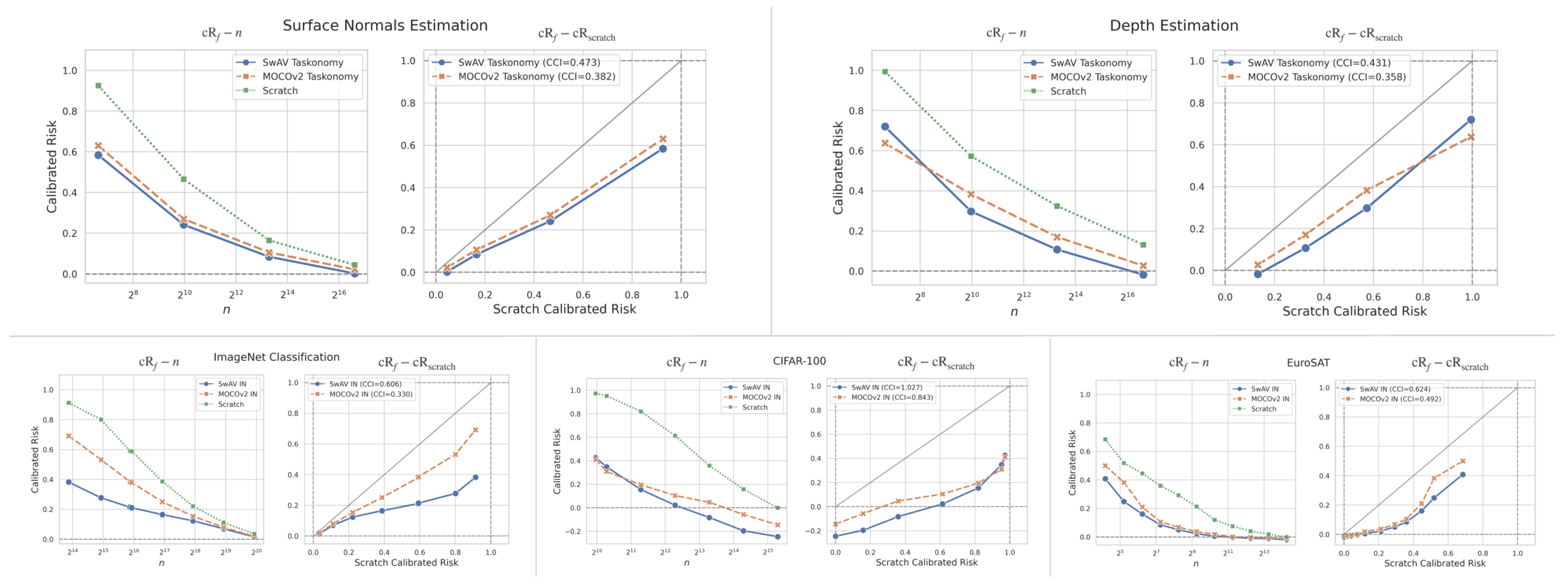}
    \vspace*{-7mm}
    \caption{\textit{By how much does contrastive self-supervised learning outperform training from scratch?}
    We plot $\lcurve$ and $\pr$ curves for SwAV and MoCov2 across different downstream tasks.
    For the three classification tasks, we use encoders pretrained on ImageNet (denoted as IN). For surface normals and depth estimation, we use the Taskonomy pre-trained versions of the same backbones. For all the tasks, contrastive SSL performs better than no pre-training by a relatively large margin, and SwAV outperforms MoCov2 in most cases. 
    }
    \label{fig:learning-pr-curves-all}
    \vspace*{-5mm}
\end{figure*}
\label{sec:analysis}

\label{sec:exp-setting}
To illustrate how the proposed control-baselines and evaluation standard can be employed, we perform an example empirical analysis of transfer learning in Section~\ref{sec:analysis}. The current section describes our experimental setup including pre-training methods, datasets, and downstream-tasks. All experiments were conducted using PyTorch \cite{paszke2017automatic}, and the associated code will be made available online.


\subsection{Datasets and Downstream Tasks}
The literature on benchmarking transfer performances of different pre-training methods commonly focuses on \textit{image classification tasks} \cite{cole2021does, kolesnikov2020big, islam2021broad}.
Our experiments similarly incorporate three examples of such classification problems, namely ImageNet \cite{deng2009imagenet}, CIFAR-100 \cite{krizhevsky2009learning}, and EuroSAT \cite{helber2019eurosat}.
In addition to classification, we also include \textit{pixel-wise regression tasks} using the Taskonomy dataset \cite{zamir2018taskonomy}.
In total, we use four different datasets, each representing a distinct image domain with different properties.


\textbf{Semantic Classification Tasks}.
ImageNet is a standard large-scale computer vision dataset of 1.3M natural images, each containing an object from one of the 1000 classes. We also use the relatively smaller CIFAR-100 and EuroSAT datasets to evaluate how well self-supervised learned representations transfer to different domains when only a small amount of labelled data is available.
For ImageNet and CIFAR-100, we use the standard train/test splits, and split the EuroSAT dataset randomly.
We train all models using the cross-entropy loss and evaluate them on the corresponding test sets using error rate.
As the blind-guess prediction, we use the most common class.
We train the maximal-supervision control baselines for CIFAR-100 and EuroSAT from scratch, using all available training images and use weights available from the PyTorch library \cite{paszke2017automatic} for ImageNet.

\textbf{Pixel-Wise Regression Tasks.}
We use two common pixel-wise regression tasks: depth and surface normals estimation from the Taskonomy dataset \cite{zamir2018taskonomy}.
Taskonomy contains 4M images of natural indoor scenes from about 600 different buildings.
We use images from the official full+ split buildings, and fix a random subset of 60K images from the set of test buildings to evaluate the final performances.
We use $L_1$-norm as the loss for training on both downstream tasks.
The blind-guess control baseline for both tasks is computed as the pixel-wise median over a set of 60K train images.
The maximal-supervision control baselines are trained from scratch using 1.2M images from the same full+ split, and reach a level of performance close to the one reported by \cite{zamir2020robust}.

\subsection{Pre-Training Methods}


\textbf{Contrastive Self-Supervised Learning.}
In our showcase analysis, we consider representative samples of contrastive self-supervised learning methods: SwAV \cite{caron2020unsupervised}, MoCov2 \cite{chen2020improved}, SimCLR \cite{chen2020simple}, and PIRL \cite{misra2020self}.
These methods have the main attributes of modern contrastive methods and most of them achieve state-of-the-art performance on standard benchmarks.



\textbf{Non-Contrastive Pre-Training.}
As examples of non-contrastive approaches, we consider two pretext tasks based on image colorization \cite{zhang2016colorful} and jigsaw puzzle \cite{noroozi2016unsupervised}.
We also use modern non-contrastive augmentation-based methods SimSiam \cite{chen2021exploring} and Barlow Twins \cite{zbontar2021barlow}.
We use the pre-trained models from the corresponding github releases{\footnote{SwAV: \url{https://github.com/facebookresearch/swav}, MoCov2: \url{https://github.com/facebookresearch/moco/}, SimSiam: \url{https://github.com/facebookresearch/simsiam}}}, as well as pre-trained models from the VISSL library { \footnote{VISSL: \url{https://github.com/facebookresearch/vissl}}} \cite{goyal2021vissl}.

\textbf{Supervised Pre-Training.} 
\label{sec:methods}
We consider two supervised tasks in our study to measure whether there is a gap between supervised and self-supervised pre-training approaches.
First, we use a standard ImageNet pre-trained encoder, as commonly employed by many evaluation studies, and generally adopted as an initialization in the community.
Second, for depth and surface normals estimation, we choose the corresponding pre-training tasks from the Taskonomy dictionary that were empirically found to result in the best transfer performance. As reported by \cite{zamir2018taskonomy}, this task is reshading for both.
We pre-train the reshading encoder on the Taskonomy dataset using the same number of images as ImageNet pre-training.
For the ImageNet encoder we use the weights provided by PyTorch.


\subsection{Architecture and Training Details}
All pre-trained encoders share the same ResNet-50 architecture \cite{he2016deep}.
For transfers to downstream tasks, we use two types of decoders.
For classification tasks, we use a single fully-connected layer that takes the output of the final encoder's layer and outputs the logits for each class.
For pixel-wise regression tasks, we use a UNet-style \cite{ronneberger2015u} decoder with six upsampling blocks and skip-connections from the encoder layers of the same spatial resolution.
We randomly sub-sample $n$ images from the corresponding training set to form each data-regime and split them into train and validation sets.
We use the validation split for early-stopping, as well as to choose the hyper-parameters for different downstream tasks.
The final performance is evaluated on the fixed test split.


\section{Empirical Analysis}
\label{sec:empirical-analysis}
The main goal of this section is to showcase an actual application of the proposed evaluation standard on a set of exemplary scenarios. It presents a targeted and small-scale empirical study that poses a set of questions about the transfer performance of self-supervised pre-training methods, and proceeds with a subsequent analysis that utilizes the suggested controls and visualizations to address them.

\begin{figure*}
    \centering
    \vspace*{-5mm}
    \includegraphics[width=\textwidth]{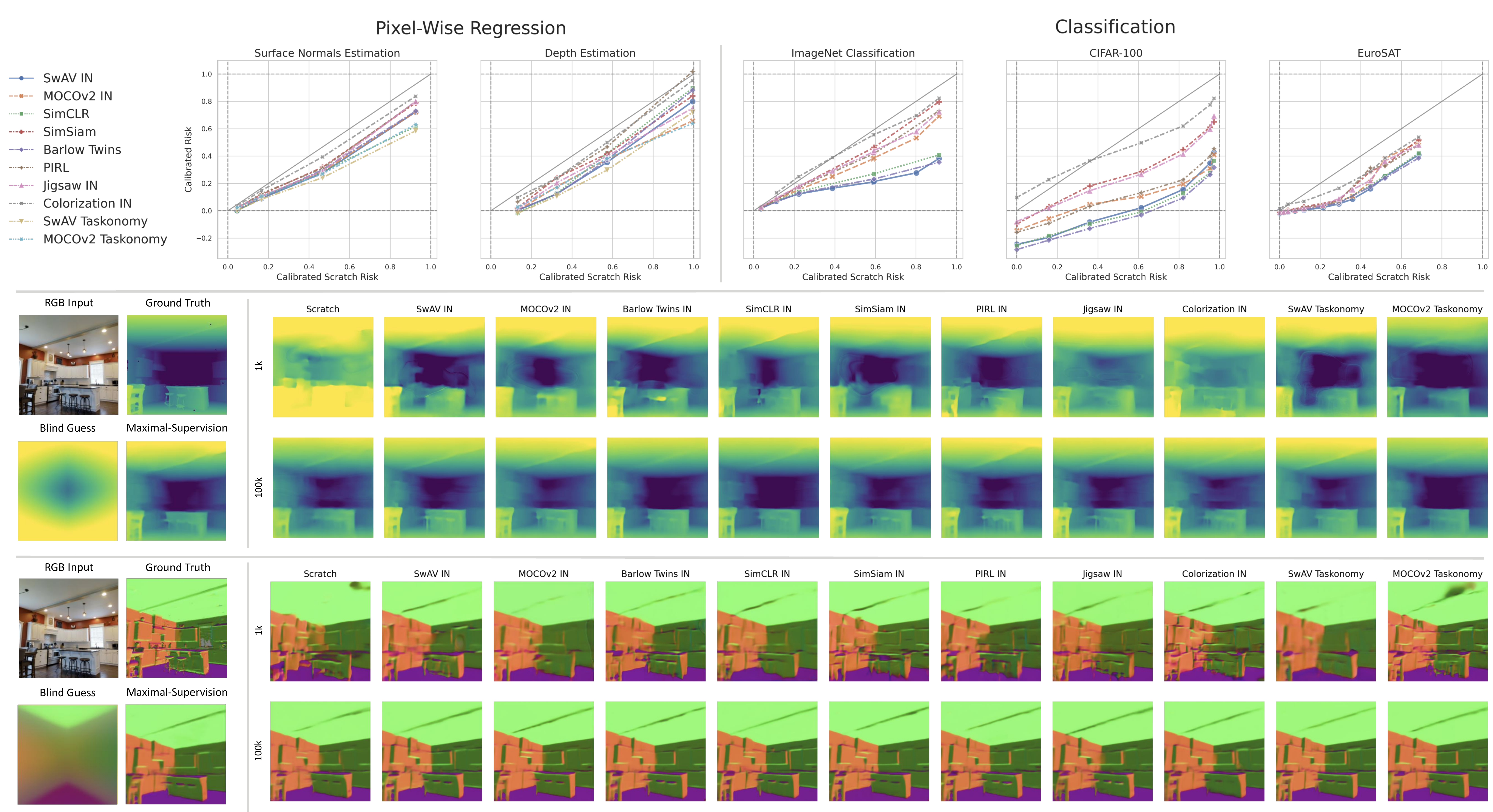}
    \vspace*{-7mm}
    \caption{
    \textit{Do different tasks benefit differently from self-supervised pre-training?}
    The $\pr$ plots above give a comparison across all pre-training methods for each downstream task. The images given below show corresponding visualizations of depth and normals predictions for different methods at two different dataset sizes.
    It can be observed that I. the differences among different pre-training methods are more pronounced for classification tasks compared to pixel-wise regression tasks, and II. the best pre-training result for classification is notably better than that of dense regression. Particularly at the high data-regime, such differences become insignificant. \textit{This suggests the development of pre-training methods may be more curated towards downstream classification tasks.}
    }
    \label{fig:pr-wr-all-tasks}
    \vspace*{-5mm}
\end{figure*}

\subsection{
By how much does contrastive self-supervised learning outperform training from scratch?
}
To address this question, we visualize the $\pr$ and $\lcurve$ curves for MoCov2 and SwAV pre-training methods.
For both methods, we use Taskonomy pre-trained encoders for pixel-wise regression tasks and ImageNet pre-trained versions for classification tasks, as we found them to perform better.
The resulting curves are shown in Fig.~\ref{fig:learning-pr-curves-all}.

We observe that both contrastive self-supervised methods outperform training from scratch on all downstream tasks in all data-regimes with a relatively large margin which, as one would expect, diminishes with more labelled data available for transfer.
Between two pre-training methods, SwAV outperforms MoCov2 in most cases.

We note that the curves for CIFAR-100 eventually fall below the x axis. These negative calibrated risk values mean that transfer learning eventually starts outperforming the maximal-supervision control baseline (i.e., training from scratch using a large amount of available data) due to the small scale of the CIFAR-100 dataset. \textit{The relative improvement and CCI computations with respect to the scratch performance, as well as comparisons between different methods for the same task are still valid in this case}, and the main difference is that the curves no longer converge to the maximal-supervision baseline since it is no longer a good approximation of the best achievable performance on the downstream task. 

\subsection{Do different tasks benefit differently from self-supervised pre-training?}
\label{sec:different-tasks}


The plots in Fig.~\ref{fig:pr-wr-all-tasks} show how $\pr$ curves for different pre-training methods compare on five different downstream tasks. We observe that the transfer performance differences between the pre-training methods are larger on classification tasks and much less pronounced on pixel-wise regression ones. Additionally, in the high data regime the differences between the methods become insignificant for pixel-wise regression tasks, which can also be qualitatively confirmed from the visualizations of depth and normals predictions at two different dataset sizes. These observations suggest that \textit{the development of pre-training methods may be implicitly biased towards classification tasks, rather than rather general purpose representations}.

\subsection{By how much do contrastive methods outperform non-contrastive methods?}

\begin{figure*}[h]
    \vspace*{-5mm}
    \centering
    \includegraphics[width=\textwidth]{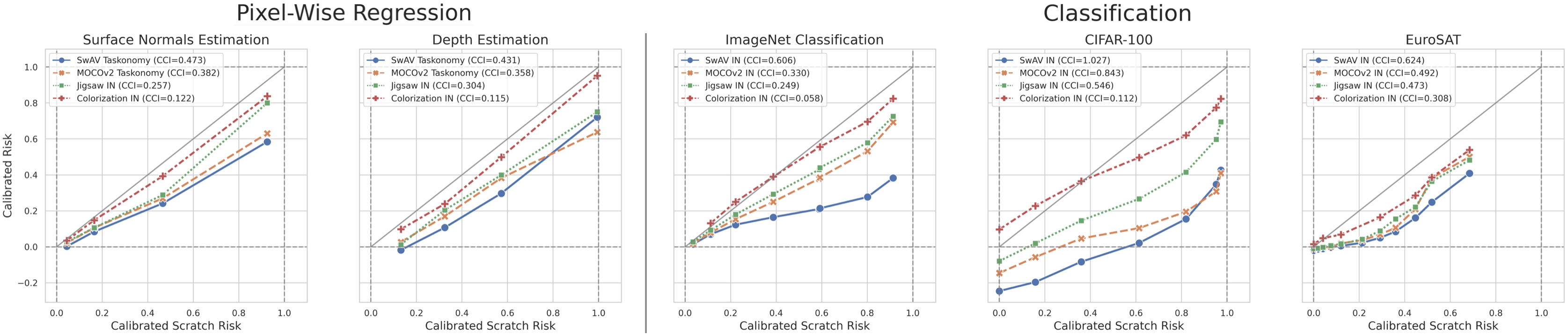}
    \vspace*{-6mm}
    \caption{
    \textit{By how much do contrastive methods outperform non-contrastive methods?}
    We plot the $\pr$ curves for contrastive pre-training methods (SwAV and MoCov2) together with non-contrastive methods (colorization and jigsaw).
    Colorization pre-training falls largely behind all other methods.
    Jigsaw pre-training, however, performs relatively similar to the two contrastive methods on pixel-wise regression tasks and EuroSAT classification, while the gap on CIFAR-100 and ImageNet is much larger (see Sec.\ref{sec:different-tasks}).
    }
    \label{fig:pr-pretext}
    \vspace*{-3mm}
\end{figure*}

\begin{figure*}
    \centering
    \includegraphics[width=0.8\textwidth]{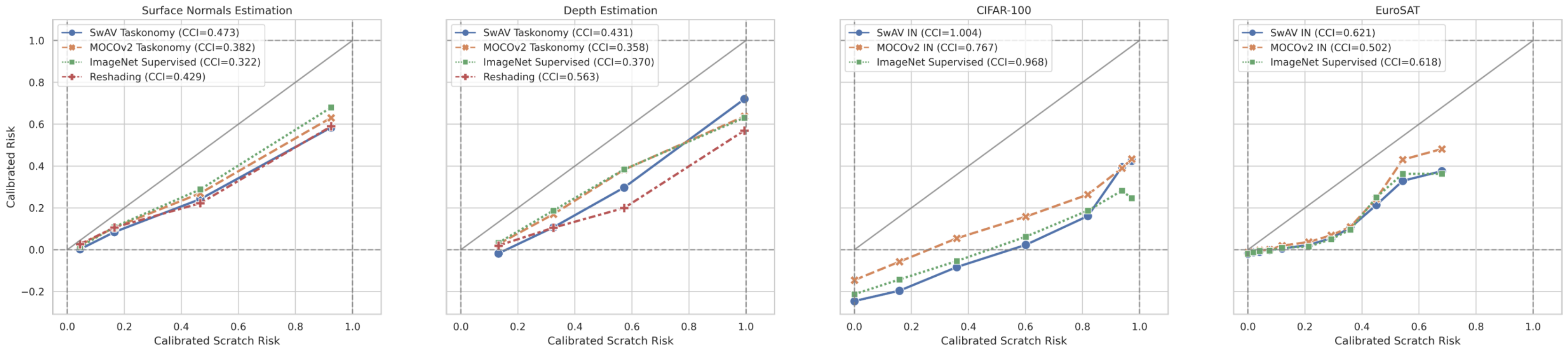}
    \vspace*{-2mm}
    \caption{
    \textit{Does self-supervised Learning outperform supervised pre-training?}
    We plot the $\pr$ curves for supervised and contrastive pre-training methods.
    ImageNet pre-training as a supervised method is included in the plots for all downstream tasks. For depth and normals estimation, we also include an encoder pre-trained on the reshading task, which is the optimal supevised transfer domain as reported by \cite{zamir2018taskonomy}.
    \textit{We observe that supervised pre-training performs similarly to contrastive methods in most cases}, except reshading supervised pre-training on depth estimation in low-data regimes where it outperforms self-supervised counterparts (discussion in Sec.\ref{sec:supervisedpretrain}). 
    }
    \label{fig:pr-supervised}
    \vspace*{-5mm}
\end{figure*}

Recent works demonstrate that for semantic classification tasks, contrastive pre-training approaches result in better transfer performance compared to non-contrastive pretext tasks \cite{he2020momentum, chen2020simple}. 
In this section, we investigate if the same trend holds for pixel-wise regression tasks.
In Fig.~\ref{fig:pr-pretext}, we show transfer performance results for colorization and jigsaw as compared to contrastive methods.
For classification tasks, we replicate the common observation that contrastive methods provide a significant improvement compared to non-contrastive methods.
However, we again observe that the difference is less pronounced on pixel-wise regression tasks, echoing the conclusion of Section~\ref{sec:different-tasks}.


\subsection{Does self-supervised pre-training outperform supervised pre-training?}
\label{sec:supervisedpretrain}
\textbf{ImageNet Supervised Pre-Training.}
In Fig.~\ref{fig:pr-supervised}, we compare ImageNet pre-training and contrastive methods.
We observe that contrastive methods generally perform comparable or better than ImageNet pre-training. However, in terms of relative improvement against the scratch control baseline, this difference was found to be relatively minor.


\textbf{Best Informed Pre-Training.}
For depth and normals estimation, we include further comparisons against the best pre-training task according to Taskonomy \cite{zamir2018taskonomy} (see Section~\ref{sec:methods} for more details).
The transfer results are presented in Fig.~\ref{fig:pr-supervised}.
We did not observe a large difference between contrastive and supervised pre-training for normals prediction, while the difference is more pronounced for depth estimation. This seems consistent with \cite{zamir2018taskonomy} where the best source tasks were found to work well \emph{if} they it is a close match with the target.

\section{Conclusion and Limitations}
We put forth an evaluation standard for measuring the effectiveness of pre-training methods for transfer learning, which incorporates three important control baselines:
\begin{itemize}
    \item the \textit{scratch} control baseline to disentangle the benefits of transfer learning from \textit{the inductive biases introduced by other training choices} that are independent from the particular pre-training method,
    \item the \textit{maximal-supervision} control baseline to account for \textit{the effects of irreducible factors inherent to the downstream-task} that upper-bound the best possible transfer performance,
    \item the \textit{blind-guess} control baseline to normalize away the benefits of \textit{the statistical regularities of particular datasets} that lower-bound the worst possible transfer performance.
\end{itemize}
In Section~\ref{sec:empirical-analysis}, we used these control baselines to define the \textit{calibrated risk} metric $\SR_f$, with the goal of framing downstream task dependent empirical risk in an interpretable scale. We further showed two visualizations to plot calibrated risk curves, with the overall goal of providing an accessible way to judge the transfer efficacy of a given pre-training method, as well as to facilitate comparisons across methods. In Section~\ref{sec:analysis}, we have provided a targeted small-scale empirical study that showcases how the proposed evaluation standard can be applied. Through this study, we mainly observed that:  
\begin{itemize}
    \item Classification tasks consistently benefit more from contrastive self-supervised pre-training, compared to non-classification tasks such as pixel-wise regression (depth and surface normals estimation).
    \item The differences between self-supervised pre-training methods (and particularly between contrastive and non-contrastive approaches) are much less pronounced for non-classification tasks, as compared to classification tasks where there is a clear improvement from using contrastive methods. These are useful indications and insights for self-supervised learning research toward developing truly more general representation. 
\end{itemize}

Our study, however, comes with several limitations:\\
\textbf{Cross-Task Comparisons.}
Comparing methods across different tasks using their calibrated risk values makes the assumption that the prediction fidelity of each task can be captured by an affine transformation of its task-specific loss function, which may not hold true in practice. \\
\textbf{Other Benefits of SSL.} 
Besides reducing the need to label data, self-supervised learning provides other benefits, such as enabling training on continuous data streams (as opposed to fixed training sets) or reducing the reliance on rigid category definitions. This paper focused on quantifying the effectiveness of self-supervised learning in terms of transfer learning and reducing labeled data demands only.\\
\textbf{Pre-training and Downstream Tasks Diversity.} 
In comparison to other works whose main focus is to provide a comprehensive empirical study \cite{kotar2021contrasting, chaves2021evaluation, goyal2019scaling, islam2021broad, van2021benchmarking}, this paper mainly aims to propose an evaluation standard, and consequently we consider a more limited set of pre-training and downstream-tasks mainly curated to showcase the importance of the proposed control baselines.\\
\textbf{The Effects of Pre-training Dataset Size.}
We evaluate the transfer efficacy as a function of the transfer dataset size, but the dependency on the pre-training dataset size is another important factor that determines the overall transfer performance, as studied in \cite{goyal2019scaling, he2020momentum}.

\section{Acknowledgment}
We thank Roman Bachmann for sharing the transfer codebase for Taskonomy and for useful discussion.

{\small
\bibliographystyle{ieee_fullname}
\bibliography{cvpr2022_reference.bib}
}

\end{document}